\documentclass[acmtog]{acmart}
\usepackage{booktabs} 
\usepackage{multirow}
\usepackage{xcolor}
\usepackage[ruled]{algorithm2e} 

\SetAlFnt{\small}
\SetAlCapFnt{\small}
\SetAlCapNameFnt{\small}
\SetAlCapHSkip{0pt}

\usepackage{xspace}

\newcommand{\etal}{{\emph{et al.}}}

\acmJournal{TOG}
\acmVolume{45}
\acmNumber{4}
\acmArticle{39}

\setcopyright{acmlicensed}

\authorsaddresses{This work has been partially supported by the National Key RD Program of China (No. 2024YFB2809102), the National Natural Science Foundation of China (No. 62125107), and the Fundamental Research Funds for the Central Universities (No. 2233100028).
\newline
Authors' Contact Information: Dixuan Lin, Beijing Normal University, China; e-mail: dixuan@seas.upenn.edu; 
Yuxiang Zhang, Tsinghua University, China; e-mail: zhangyuxthu@qq.com;
Mengcheng Li, Tsinghua University, China; e-mail: li-mc18@mails.tsinghua.edu.cn;
Wei Jing, Lenovo, China; e-mail: jingwei1@lenovo.com;
Qi Yan, Lenovo, China; e-mail: yanqi4@lenovo.com;
Qianying Wang, Lenovo, China; e-mail: wangqya@lenovo.com;
Yebin Liu, Tsinghua University, China; e-mail: liuyebin@mail.tsinghua.edu.cn;
Hongwen Zhang (corresponding author), Beijing Normal University, China; e-mail: zhanghongwen@bnu.edu.cn;
}

\begin{document}
\title{OmniHands: Robust Motion Capture of Interactive Hands via A Versatile Transformer}

\author{Dixuan Lin}
\affiliation{%
 \institution{Beijing Normal University}
 \country{China}}
 \author{Yuxiang Zhang}
\affiliation{%
 \institution{Tsinghua University}
 \country{China}}
 \author{Mengcheng Li}
\affiliation{%
 \institution{Tsinghua University}
 \country{China}}
 \author{Wei Jing}
\affiliation{%
 \institution{Lenovo}
 \country{China}}
 \author{Qi Yan}
\affiliation{%
 \institution{Lenovo}
 \country{China}}
 \author{Qianying Wang}
\affiliation{%
 \institution{Lenovo}
 \country{China}}
  \author{Yebin Liu}
\affiliation{%
 \institution{Tsinghua University}
 \country{China}}
  \author{Hongwen Zhang}
  \authornote{Corresponding author.}
\affiliation{%
 \institution{Beijing Normal University}
 \country{China}}


\begin{abstract}

In this paper, we introduce OmniHands, a universal approach to recovering interactive hand motions and their relative movement from monocular or multi-view inputs. Our approach addresses three major limitations of previous methods: lacking a unified solution for handling single hand and interactive hands image inputs, lacking a versatile framework to support various 4D tasks such as video-based tasks and multi-view tasks, and unable to maintain robust performance in real-world data. To overcome these challenges, we develop a universal architecture  with novel tokenization and contextual feature fusion strategies, capable of adapting to a variety of tasks. Specifically, we propose a Relation-aware Two-Hand Tokenization (RAT) method to embed positional relation information into the hand tokens. In this way, our network can handle both single-hand and two-hand inputs and explicitly leverage relative hand positions, facilitating the reconstruction of intricate hand interactions in real-world scenarios. As such tokenization indicates the relative relationship of two hands, it also supports more effective feature fusion. To this end, we further develop a 4D Interaction Reasoning (FIR) module to fuse hand tokens in 4D with attention and decode them into 3D hand meshes and relative temporal movements. The efficacy of our approach is validated on several benchmark datasets. The results on in-the-wild videos and real-world scenarios demonstrate the superior performances of our approach for interactive hand reconstruction. The code and video results can be found in \href{https://omnihand.github.io/}{omnihands.github.io}.

\end{abstract}

%
%


%
%


\begin{teaserfigure}
    \centering
    \includegraphics[width=1.0\linewidth]{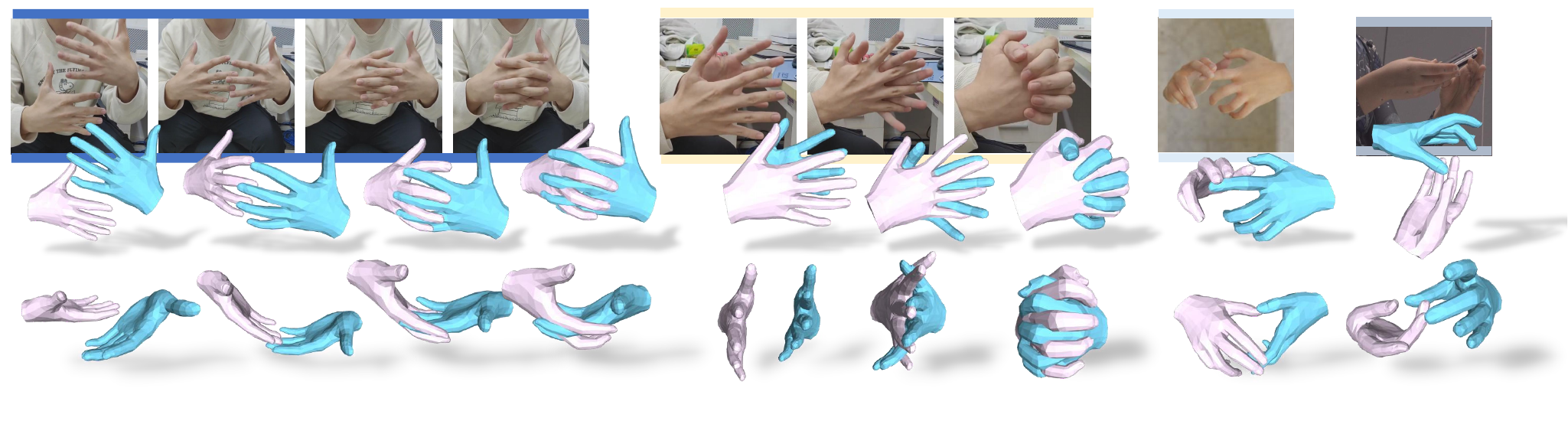}
    \vspace{-9mm}
    \caption{The proposed method, OmniHands, can robustly recover interactive hand motions and their relative movement from monocular inputs.
    }
    \label{fig:teaser}
\end{teaserfigure}

\maketitle

\section{Introduction}
Hand mesh recovery is a fundamental technology for numerous advanced applications like virtual/augmented reality, human-computer interaction, robotics, and embodied AI. 

Despite the remarkable progress achieved by existing methods~\cite{intaghand, lin2021metro, acr} in this field, three major challenges remain unresolved. First, there is a lack of a universal method that can handle various input statuses for both single-handed and two-handed input. When using existing single-hand reconstruction models to reconstruct two hands, while they can individually reconstruct each hand, they fail to address the issue of hand occlusion and cannot compute the relative distance between the hands. On the other hand, current two-hand reconstruction models, due to their use of single images containing both hands as input, are limited to reconstructing scenarios where the hands are in close proximity, thereby exhibiting poor generalizability. Second, current hand recovery methods address specific subproblems such as monocular hand reconstruction, video-based reconstruction, and multi-view reconstruction, however, a unified and general model to handle these tasks is still lacking. Third, although current hand reconstruction methods perform well on test data, their effectiveness in in-the-wild scenarios is limited due to the inadequate size of training datasets, resulting in poor practical value.

The success of recent approaches in transformer-based large-scale models such as GPT-4\cite{achiam2023gpt}, ViTpose \cite{vitpose}, HaMeR \cite{hamer} provided us with the insight to investigate the usage of simple and high-capacity models along with large-scale in-the-wild data. Therefore, we proposed OmniHands, a transformer-based universal solution to recover interactive hand meshes cross various scenarios.
The transformer possesses versatility, enabling it to perform unified forward operations on different input forms such as single-frame input, temporal input, and multi-view input. Additionally, multi-layer transformers' ability to process large datasets allows us to train a generalized model that remains robust in real-world scenarios across various lighting conditions, backgrounds, and interaction poses.

In the proposed method OmniHands, we unified the following aspects to achieve robust 4D hand mesh recovery:
\textbf{i) Status of Hand Images.} 
The input can be single hand images or interactive hands images. In previous inter-hand recovery methods, a single image containing both hands is typically used as input. However, to accommodate both single-hand and inter-hand recovery scenarios, we need to unify the input format that handles both types of input while preserving the relative positions of the hands. 
\textbf{ii) View Points.}
Inputting sequences of images from multiple camera perspectives captured at the same moment to enhance details through information supplementation across different viewpoints. 
\textbf{iii) Temporal Sequences.}
Inputting a monocular dynamic sequence, make more accurate predictions by evaluating hand movements under conditions of motion blur and object occlusion. Since the multi-view data and temporal data can both be seen as contextual sequences, a unified module can be applied to fuse the contextual information.


To address these issues, we proposed a Relation-aware Two-Hand Tokenization (RAT) method and a 4D Interaction Reasoning (FIR) module to unifiedly process features in spatial context or temporal context.

In Relation-aware Two-Hand Tokenization (RAT), we crop and tokenize the RGB information of each hand individually, while maintaining hand-hand relative position information in tokens.
Such a tokenization method brings the following benefits threefold.
i) RAT is not reliant on the co-occurrence of two hands and can thus handle both single-hand and two-hand inputs in a unified manner.
ii) RAT takes the overlapped status of two hands into account and adjusts the tokens accordingly, which allows the network to learn a more concentrated distribution of the hand tokens regardless of the various scales and positions.
iii) RAT explicitly encodes the relative positions of two hands into the tokens, facilitating the reconstruction of close and intricate interaction in real-world scenarios.

Upon obtaining the image tokens, we integrate a Context Interaction Reasoning (FIR) module to unifiedly fuse contextual relations within the context channel. These contextual relations could be temporal (over time) or spatial (from multiple camera views).
Thanks to the proposed tokenization method RAT, FIR can fuse both the contextual information more effectively as the hand tokens indicate the relative relationship of two hands, thereby enhancing the robustness of hand mesh and contextual movement recovery.

In summary, the main contributions of this work can be listed as follows:
\begin{itemize}
\item We introduce a Relation-aware Two-Hand Tokenization (RAT) in our transformer. By embedding positional relation information in the tokenization process, RAT enables a generic and informative process of hand inputs across diverse scenarios, regardless of whether the two hands are interacting or separated.
\item We propose a 4D Interaction Reasoning (FIR) module to fuse spatial or temporal hand features. The collaboration of RAT and FIR further enhances the stability of hand mesh and temporal movement recovery.
\item Our final solution, OmniHands, achieves robust hand motion capture in real-world scenarios, serving as a versatile solution to handle hand inputs in various forms, whether the images contain single or two hands, in single frames or temporal sequences or multi-view sequences.
\end{itemize}

\section{Related Work}

\subsection{Monocular Hand Motion Capture}
Recent years have witnessed great progress in hand mesh recovery from monocular RGB input.
Following the advances in monocular human mesh recovery~\cite{kanazawa2018end,kolotouros2019graphcmr,kolotouros2019learning,zhang2020densepose2smpl,zhang2021pymaf,tian2023recovering}, some researchers~\cite{Baek_2019_CVPR,hamer,Park_2022_CVPR_HandOccNet,Zhang_2019_ICCV,boukhayma20193d} use neural networks to regress the driving parameters of hand templates such as MANO~\cite{MANO:SIGGRAPHASIA:2017} to recover hand meshes, while others prefer to directly estimate the vertex coordinates of hand meshes through GCN~\cite{Ge_2019_CVPR,2020Weakly,tang2021towards,Choi_2020_ECCV_Pose2Mesh,bib:CMR,bib:MobRecon} or transformer~\cite{meshgraphormer,lin2021metro}.

Recently, there have also been some efforts to simultaneously recover both two hands of humans. However, two hands mesh recovery is much more challenging than one hand, due to larger pose distribution and more severe occlusion. Furthermore, compared to the large-scale single-hand datasets~\cite{hamer} with rich capture scenarios and postures, existing two-hand datasets are much fewer, and most of them are captured in limited environments.

A straightforward method for two-hand mesh recovery is to locate and crop each hand separately from the input image, and then convert the task into single-hand scenarios by processing two cropped hand images individually. This strategy is widely used in full-body pose estimation frameworks~\cite{pymafx2023,Joo_2018_CVPR,ExPose:2020,rong2020frankmocap,PIXIE:2021,Moon_2022_CVPRW_Hand4Whole,lightcap2021}. However, this approach is still based on single-hand recovery network, which usually performs poorly under close interaction situations. Moreover, the relative translation between two hands is ambiguous when the two hands are processed separately.

\subsection{Hand Motion Capture under Interaction}
To better address two-hands pose estimation and mesh recovery in close interaction scenarios, some researchers~\cite{intaghand,ith3d,acr,reconprior,kim2021end,Moon_2020_ECCV_InterHand2.6M,meng2022hdr,Li2024HHMR} used one image containing both hands as input. Zhang \etal~\cite{ith3d} utilized 2d joints projection attention and context-aware refinement module
to improve the pose accuracy. {Li \etal~\cite{intaghand}, Cho \etal~\cite{cho2023transformer} and Jiang \etal~\cite{jiang2023a2j} utilize transformer module~\cite{vit} to learn implicit attention between two hands.} Zuo \etal~\cite{reconprior} propose a data-based prior of two interacting hands. Yu \etal~\cite{acr} constructed the attention relationship between the two hands based on 2D hand center maps.
Although the above approaches have produced promising results, there are still some limitations. 
Because these methods use a single image containing both hands as input, when the two hands are far apart, the area proportion of hands in the input image will be too small, which can reduce the accuracy of mesh recovery.  {Moon \etal~\cite{inthewild} separates the two hands as input, but its prediction of the relative position between the hands relies on bounding box information rather than visual cues from the image. Moreover, it also requires input images that contain both hands since the separation is conducted during forward process. Furthermore, these above network designs inherently lack the ability to recover single-hand mesh and only can operate under two-hand scenarios.}

Therefore, we propose a new network that is compatible with both single-hand recovery and two-hand recovery. We take one or two cropped hand images as input, enabling utilize large-scale single-hand datasets to boost two-hand recovery results. We also tokenize the 2D relative distance and overlap between two cropped hand images by a powerful transformer~\cite{Attention_is_all_you_need,vit} module, resulting in accurate relative root translation estimation and interacting hands mesh recovery.

\subsection{Temporal Hand Motion Capture}

Several methods have been proposed for hand mesh recovery from temporal sequence data~\cite{liu2021semi,hasson20_handobjectconsist,seqhand,deformer,vibe,tcmr}.  {RGB2Hands~\cite{wang2020rgb2hands} proposed a real-time hand tracking approach that utilizes both segmentation and depth prediction to enhance performance.} Hasson \etal~\cite{hasson20_handobjectconsist} leverage the optical flow of nearby frames to constrain the motion of hands and objects by photometric supervision. SeqHAND~\cite{seqhand} incorporates
the convolution-LSTM layer to capture the sequential relationship between consecutive hand poses. Deformer~\cite{deformer} proposes a novel dynamic fusion module that explicitly deforms nearby frames with clear hand visibility for robust hand pose estimation from temporal input. VIBE~\cite{vibe} and TCMR~\cite{tcmr} utilize the recurrent neural network to produce temporally consistent motion.

However, these methods rely solely on sequence data. Our approach introduces a 4D Interaction Reasoning (FIR) module, enabling our method to not only recover accurate two-hand interacting meshes from single-frame input but also produce smooth and stable 4D hand recovery from occluded and motion-blurred temporal input.

\subsection{Multi-View Hand Motion Capture}
Recovering hands from multi-view input is a similar task as multi-view human reconstruction. Many recent researches have been proposed for multi-view 3D human pose estimation ~\cite{qiu2019cross, dong2019fast,tu2020voxelpose,pavlakos2017harvesting,zhang2021adafuse}. Some of the methods~\cite{qiu2019cross,dong2019fast,tu2020voxelpose} focus on matching the projection of 3D key-points with their corresponding 2D key-points from multiple camera views, and some~\cite{pavlakos2017harvesting,zhang2021adafuse} focus on fusing the 2D visual features to build a 3D structure.

Recent research on multi-view hand reconstruction has seen various innovative approaches. Some researchers \cite{corona2022lisa, guo2023handnerf} utilize implicit fields to recover the shape and appearance of hands. Yang \etal \cite{yang2023poem} employ hand vertices as a point cloud embedded in the intersection volume of frustums from multiple cameras to integrate multi-view information. Wang \etal \cite{wang2024sima} adapt multi-view hand data to enhance the performance of single-view hand reconstruction. Additionally, Gan \etal \cite{gan2024fine} propose a neural rendering scheme to refine both the hand mesh and textures. However, most multi-view methods require extrinsic and intrinsic calibration information for each camera to accurately determine the 3D positions of hands. In contrast, our approach is capable of utilizing only visual information to recover precise hand meshes.

\section{Method}


\begin{figure*}[t]
	\centering
	\includegraphics[width=1.0\textwidth]{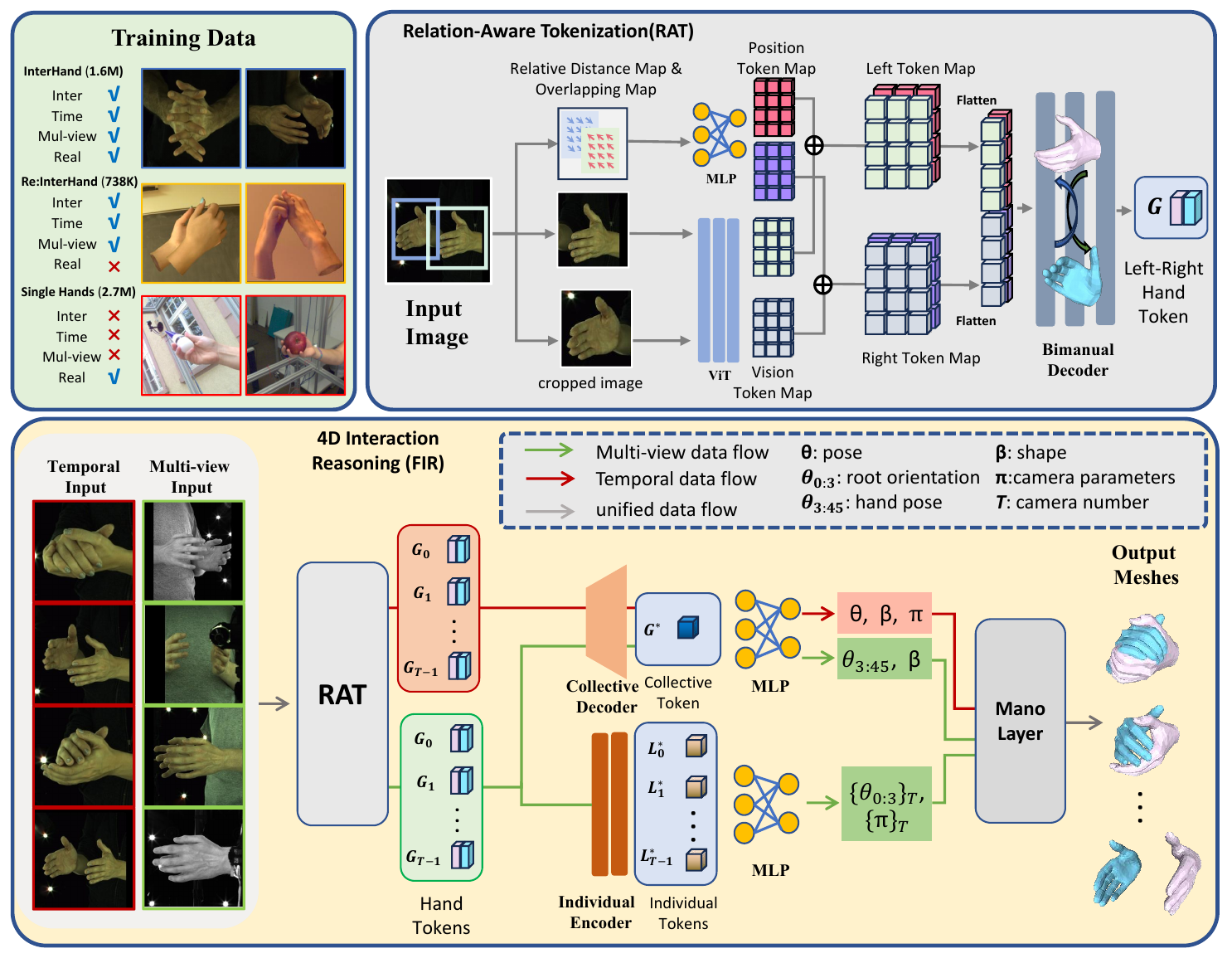}
        
	\vspace{-3mm}
    \caption{Overview of our OmniHands framework. OmniHands is a transformer-based network, which takes various forms of inputs and estimates two-hand meshes with their relative positions. General modules are used to process different forms of input data. For simplicity of representation, since the hand tokens of both hands will undergo the same processing in FIR, the hand tokens $G^*, L^*$ in the diagram represent those of one hand. In the top left corner we present our training set, indicating the data volume and whether including Interaction, Time sequences, Multi-view, Real-World data.
    }
    \Description{A Figure.}
	\label{fig_main}
\end{figure*}

In this Section, we present the technical details of OmniHands. 
As illustrated in Figure \ref{fig_main}, OmniHands is a transformer-based network, which can accept multiple forms of input: single-frame images, videos, and multi-view images to estimates the hand meshes, as well as their relative position. 
In the proposed transformer, the key components include 
Relation-aware Two-hand Tokenization (RAT) and 4D Interaction Reasoning (FIR).
As will be discussed shortly, the tokenization and reasoning strategies make OmniHands a generic solution to handle hand images in various forms, regardless of the images with single or two hands, in a single frame or temproal sequences or multi-view sequences.


\subsection{Problem Formulation}
Given a single-frame or a temporal sequence or a multi-view sequence of images $I$, our goal is to reconstruct the hand meshes and their relative position.
We leverage the widely used parametric hand model MANO~\cite{MANO:SIGGRAPHASIA:2017}. MANO defines a function $\mathbf{Mn}(\theta,\beta)$ that takes two parameters pose $\theta \in \mathbb{R}^{48}$ and shape $\beta \in \mathbb{R}^{10}$ as input, returns 3D hand mesh $V \in \mathbb{R}^{778\times 3}$ and 3D joint keypoints $V \in \mathbb{R}^{21\times 3}$. The outputs of our model are MANO parameters of two hands: $\theta_R, \theta_L, \beta_R, \beta_L$, by which we can regress meshes and joints of each hand, and a vector $\upsilon$ to represent the joint root distance of two hands. We define $\upsilon$ to be the location of the left root joint location when we subtract the joints of both hands from the right hand's 3D root joint location.
Note that our method can robustly handle both single-hand and two-hand inputs.
When there is only a single hand visible in the images, our method is simply degraded to the estimation of the mesh of one hand.

\subsection{Relation-aware Two-Hand Tokenization}

Existing approaches to hand mesh recovery typically take the cropped hand images as input, but neglect the hand relationship in the images.
Besides, these approaches handle two-hand and single-hand scenarios in different manners, resulting in jittering artifacts when the two hands operate in interaction and separately.
To handle these issues in our transformer-based architecture, we propose a Relation-aware Two-Hand Tokenization (RAT) method to take the relation of two hands into consideration and handle inputs in various scenarios.
Following recent state-of-the-art practices in 3D human/hand mesh recovery~\cite{vitpose,4dhuman,hamer}, we utilize ViT~\cite{vit} as the visual encoder to tokenize image patches.
In contrast to these methods, we handle both the single-hand and two-hand inputs in a unified manner and consider embedding positional relation information of each hand into its tokens during the tokenization process.

To achieve this, we first locate each hand from inputs and center it into the corresponding left and right hand images.
This strategy allows our method to learn a more concentrated distribution of the hand tokens regardless of the various scales and relative positions.
Moreover, by tokenizing the images of two hands individually, our method can maintain the stability of inputs across various interaction scenarios, eventually enhancing the robustness of our solution.
To compensate for the lack of relation information in the separated tokens, we further consider enriching the hand tokens with relative positional embedding information so that they can be aware of the relative position with respect to each hand and the input images.

Specifically, for each input image, we follow ViT~\cite{vit} to segment it into $H\times W$ image patches and subsequently feed them into the ViT backbone, resulting in a vision token map $M \in \mathbb{R}^{H\times W\times C}$ .
To enrich each token map with relationship information, we further calculate relation maps $\{\mathcal{R}_{R2L}, \mathcal{R}_{L2R}\}$ and embed them into the final image tokens, so that they can contain both the patch-wise visual knowledge and patch-to-patch relation information.
Basically, the relation maps are the composition of the relative distance map and the overlapping map.
Without losing generalization, we describe the right-hand relation map $\mathcal{R}_{R2L}$ in detail as follows.
The left-hand relation map $\mathcal{R}_{L2R}$ can be calculated similarly.

\paragraph{Relative Distance Map}
The relative distance map is calculated based on the hand position in the original images.

 {Let $\mathcal{B}=[c_x, c_y, s_x, s_y]$ be the bounding box of one hand, $c_x, c_y$ the x-coordinate and y-coordinate of the center position, and $s_x, s_y$ the box scale lengths on x-axis and y-axis.}
We obtain a position map $\mathcal{C}\in\mathbb{R}^{H\times W \times 2}$, representing positions of the center in each image patch. For the patch $p(i,k)$ at the $i$-th column and the $k$-th row, its value is
\begin{equation}
    \label{eq_center_map}
        \mathcal{C}(p) = [c_x + \frac{(2i - W)\cdot s_x}{2 W}, c_y + \frac{(2k - H)\cdot s_y }{2 H}].
\end{equation}
In our method, for each right hand and left hand, we first obtain their position maps $\mathcal{C}_R, \mathcal{C}_L$ and then calculate a relative distance map $\mathcal{D}\in \mathbb{R}^{H\times W \times 2}$. The relative distance token at $p(i,k)$ of the right hand is formulated as
\begin{equation}
    \label{eq_dis_map}
        \mathcal{D}_{R2L}(p) = \psi(\tau \cdot \frac{\mathcal{C}_R(p) - \mathcal{C}_L(p)}{s}) 
\end{equation}
where $s=(s_x,s_y)$ is the scale of right-hand bounding box, $\psi$ is an activation function, $\tau$ is a hyper-parameter used to adjust the activation distance.
In this way, the relative distance map is activated when two hands are close.
In our experiments, we use Sigmoid as the activation function $\psi$.


\paragraph{Overlapping Map}
To explicitly represent the overlapped relation of two hands, we further introduce the overlapping map. Denoting $Inside(p(i,k), \mathcal{B})$ as a function that indicates whether the center of patch $p$ is inside the bounding box $\mathcal{B}$, we formulate the the overlapping map $\mathcal{O} \in \mathbb{R}^{H\times W\times 1}$ as
\begin{equation}
    \label{eq_over_map}
    \mathcal{O}_{R2L}(p) = \begin{cases}
        1, &  Inside(p, \mathcal{B}_L) \\
        -1, & \text{otherwise}
    \end{cases}
\end{equation}

Given the above relative distance map and overlapping map, the relation map is calculated by first concatenating them in the last dimension and using an MLP to match the dimension with the image feature dimension $C$:
\begin{equation}
    \label{eq_rel_map}
    \mathcal{R}_{R2L} = MLP(\mathcal{D}_{R2L}\oplus\mathcal{D}_{L2R}\oplus\mathcal{O}_{R2L}),
\end{equation}
where $\oplus$ denotes the concatenation operation.
Finally, the relation map acts as the relative positional embedding and enhances the original ViT tokens $M_{R}$ by concatenating them as the relation-aware tokens $\mathcal{M}_R \in \mathbb{R}^{HW\times C}$.

\begin{equation}
    \label{eq_rel_map2}
    \mathcal{M_R} = Flatten(M_R + \mathcal{R}_{R2L})
\end{equation}

\paragraph{Cross-hand Feature Fusion}
With the token maps, we can fuse the cross-hand features by a Bimanual Decoder.
Specifically, for each single frame we concatenate the tokens of two hands to $\mathcal{M} = \mathcal{M_R}\oplus\mathcal{M_L}$.
Then $\mathcal{M}$ is further fed into attention modules to fuse inter-hand information.
To enhance the adaptability of our model, we first employ Spatio Encoder to proceed with the token sequence $\mathcal{M}$:
\begin{equation}
    \label{eq_inter_attn}
    \mathcal{M^*} = FFN(Softmax(\frac{Q(\mathcal{M})K(\mathcal{M})}{\sqrt{dim(\mathcal{M})})} )V(\mathcal{M}))
\end{equation}
where $Q, K, V$  respectively represent the left-multiplied query, key, and value matrices.

As the relation-aware tokens are embedded with both visual and relation features, the self-attention module can adjust its weight adaptively based on visual and relative positional associativity and mitigate the interdependence of bilateral hand information.
For instance, when relation-aware tokens indicate that the two hands are far away from each other, the self-attention can mainly interact with the tokens of each hand individually.
The outputs $\mathcal{M}^*$ from the self-attention module are further fed into Spatio Decoder, where we take a constant vector $\alpha\in\mathbb{R}^C$ as query vector, resulting in a global hand token $G\in \mathbb{R}^{C}$ corresponding to each hand:
\begin{equation}
    \label{eq_dec_attn}
    G = FFN(Softmax(\frac{Q(\alpha)K(\mathcal{M}^*)}{\sqrt{dim(\mathcal{M}^*)})} )V(\mathcal{M}^*))
\end{equation} 
 {The vector $\alpha$ is predefined and has different value for left and right hands.} Then the hand tokens are taken for further feature fusion according to the specific task demands.

\subsection{4D Interaction Reasoning}
Regardless of whether the input consists of monocular temporal frames or multi-view non-temporal frames, we can consider the set of inputs as contextually relative, therefore employ similar methods to correlate the information contained within them. Consequently, we aim to utilize a unified module 4D Interaction Reasoning (FIR) module in order to fuse the contextual knowledge to infer the hand poses and the relative position between two hands. We will discuss the processing differences specific to temporal inputs and multi-view inputs.

\paragraph{Temporal Context Fusion}
When dealing with the temporal sequential inputs, we use $\{G_t\}_{t=1}^T$ to represent a temporal sequence of one of the two hands, for simplicity, and $\{G_t\}_{t=1}^T$ will be further proceeded by Global Context Decoder.

By fusing the sequential information, our model can predict the hand poses of intermediate frames more robustly in handling disturbances caused by object occlusions and dynamic blurring.
Note that our method is also compatible with single-frame inputs.
In this case, we stack the hand tokens repetitively to obtain the sequential token $\mathcal{G}\in \mathbb{R}^{T\times C}$. Since the temporal input has a sequential order, we need to add position embedding to $\mathcal{G}$ before inputting it into Collective Decoder to get global feature. We only output the results of the middle $\frac{t}{2}$-th frame because, when predicting the middle frame, both past and future information is available, leading to more accurate predictions:
\begin{equation}
    \label{eq_dec_attn2}
    G^* = FFN(Softmax(\frac{Q(\alpha)K(\mathcal{G})}{\sqrt{dim(\mathcal{G})})} )V(\mathcal{G})
\end{equation}


After the feature fusion of both spatial and temporal dimensions, we can obtain the integrated features $G^*_R, G^*_L \in \mathbb{R}^{C}$ as right hand collective token and left hand collective token, respectively.
Finally, these features are proceeded by a Global Regressor, an MLP-based regression network to estimate the MANO parameters $\theta_R,\theta_L, \beta_R, \beta_L$ of two hands and their relative position $\upsilon$,  {and camera parameters $\pi$ for 2D projection}.

\paragraph{Multi-view Context Fusion}
When inputting frames of multiple camera views, we get hand tokens $\mathcal{G}\in \mathbb{R}^{T\times C}$ in the same way as mentioned above, where $T$ is the number of cameras instead of time length. Then concatenating them to obtain the sequential global feature $\mathcal{G}$. Different with Temporal Context Fusion, since the camera parameters are not required as input, there isn't a fixed order of input frames and can be combined in any arbitrary sequence, therefore we don't add position embedding to $\mathcal{G}$ in this case. In temporal context fusion, we utilize information from preceding and succeeding frames to predict an accurate result of the intermediate frame. However, in multi-view context fusion, since there is no sequential order, the distance between frames is equivalent for all, enabling the mutual supplementation of information from each frame. Consequently, we can output predictions for all frames, with the collective information contributing to a precise result:

\begin{equation}
    \label{eq_dec_attn3}
    \mathcal{L}^* = FFN(Softmax(\frac{Q(\mathcal{G})K(\mathcal{G})}{\sqrt{dim(\mathcal{G})})} )V(\mathcal{G})
\end{equation}

We get individual tokens $\mathcal{L}^* \in \mathbb{R}^{T\times C}$ from Individual Encoder (Eq \ref{eq_dec_attn3}) and collective features $G^* \in \mathbb{R}^{\times C}$ from Collective Decoder (Eq \ref{eq_dec_attn2}) after fusion. Given that all frames are taken at the same time from different views, the gesture pose $\theta_{3:45}$ and shape $\beta$ should be the same, while the root orientation $\theta_{0:3}$, camera parameters $\pi$ and root offset $\upsilon$ varying. So we feed global feature $G^*$ into Collective Regressor, to regress the consistent parameters $\theta_{3:45}$ and shape $\beta$, and $\mathcal{L}^*$ into MLP-based Regressor to regress the inconsistent parameters $\theta_{0:3}$, $\pi$, $\upsilon$ for each frame.

Although we predict the root orientations and offset vectors for each frame individually, the relative orientation and distance between the right hand and left hand should remain invariant across all frames. This means that when we transform all right hands to the zero position and zero rotation, the positions and orientations of the left hands should remain consistent. Therefore, we apply a constraint (Eq \ref{eq_loss_consis}) to maintain this invariance.  {During inference, we average the relative translations and relative
orientations to ensure consistency across views.}



\subsection{Losses}
As our method predicts two hands in a disentangled way, we introduce single-hand losses and inter-hand losses to improve the accuracy of single-hand reconstruction and two-hand interaction.
\subsubsection{maxMSE Loss}
\ 
\newline
Following~\cite{deformer}, we use the maxMSE Loss to adjust the weights of each joint to maximize the MSE loss. Given joints locations $\mathcal{P^*},\mathcal{P}$, the maxMSE loss is formulated as  
\begin{equation}
    \label{eq_loss_maxmse}
    maxMSE(\mathcal{P}, \mathcal{P^*}) = \frac{\sum_{i=1}^{N}||\mathcal{P}_i - \mathcal{P^*}_i||^4}{\sum_{i=1}^{N}||\mathcal{P}_i - \mathcal{P^*}_i||^2}.
\end{equation}
With maxMSE loss, the joints or vertices with large errors, typically the vertices of fingertips, will be assigned larger loss weights.

\subsubsection{Single-Hand Losses}
\ 
\newline
 {For single-hand alignment, we train our model with MANO parameter losses, 3D losses, and 2D losses. For a single hand image}, our model predicts its pose $\theta$, shape$\beta$ and camera parameter $\pi$. With ground truth MANO parameters $\tilde{\theta}$ and shape$\tilde{\beta}$, we optimize the MANO parameters using MSE loss:
\begin{equation}
    \label{eq_loss_mano}
    L_{mano} = ||\tilde{\theta} -  \theta||^2 + ||\tilde{\beta} - \beta||^2
\end{equation}
 {With the pose and shape parameters, we can obtain 3D vertices $V$ and 3D joint keypoints $J$. With accurate 3D ground truth $\tilde{V}$ and $\tilde{J}$, we can supervise the output 3D keypoints with 3D} loss using maxMSE
\begin{equation}
    \label{eq_loss_3d}
    L_{3D} = maxMSE(\tilde{V}, V) + maxMSE(\tilde{J}, J)
\end{equation}

 {With ground truth 3D joint keypoints and camera parameters, we can get 2D keypoints $\tilde{j} = \tilde{\pi}(\tilde{J})$ on the original image. The predicted  {2D} keypoints are calculated with predicted camera parameter $\pi$. The 2D projection loss is formulated}
\begin{equation}
    \label{eq_loss_2d}
    L_{2D} = maxMSE(\pi(J), \tilde{j})
\end{equation}

\subsubsection{Inter-Hand Losses}
\ 
\newline
We utilize Inter-Hand losses in order to optimize the finger touching details of two hands. To calculate the relative distance between right hand keypoints and left hand keypoints, we relocate all keypoints to set the palm joint of right hand on the original coordinate. We denote $\phi_j(i,k)\in \mathbb{R}^{3}$ as the distance vector from the $i-th$ right hand joint to the $k-th$ left hand joint, and $\phi_v(i,k)\in \mathbb{R}^{3}$ as the distance vector of vertices, defined in the same way.

To supervise the interacting gesture, for ground truth joints and predicted joints, we calculate the distance from each right-hand joint to each left-hand joint, respectively. We formulate the joint-relation loss as:
\begin{equation}
    \label{eq_loss_joint_rel}
    L_{jrel} = \sum_{i=1}^{N_j} \sum_{k=1}^{N_j} ||\tilde{\phi}_j(i,k) - \phi_j(i,k)||^2
\end{equation}
where $N_j = 21$ represents the number of joint keypoints.
For a finer reconstruction in contact of hands, we supervise the close vertices distances as:
\begin{equation}
    \label{eq_loss_close}
    L_{close} = \sum_{i=1}^{N_v} \sum_{k=1}^{N_v} 
    \begin{cases}
        ||\tilde{\phi}_v(i,k) - \phi_v(i,k)||^{2}, & \text{if } ||\tilde{\phi}_v(i,k)||^2 \leq \alpha \\
        0, & \text{if } ||\tilde{\phi}_v(i,k)||^2 > \alpha
    \end{cases}
\end{equation}
where $N_v=778$ is the number of vertices, and $\alpha$ is a hyper-parameter, representing the distance threshold. We set $\alpha=0.005$ in experiments.  

\subsubsection{Sequential Consistency Losses}
\ 
\newline

When using video as input, we use a smooth loss function to maintain the smoothness of the predicted mesh between video frames and reduce jitter:
\begin{equation}
    \label{eq_loss_smooth}
    L_{consis} = \sum_{t=1}^{N_t-1} ||\theta_{t+1} - \theta_{t}||^2
\end{equation}
where $N_t$ is the length of video sequence.  {We choose to constrain joint velocity rather than acceleration in the smoothness loss because acceleration involves second-order differences, which empirically yield weaker gradients and often lead to less stable or effective optimization during training.}

When taking multi-view images as input, the inter-hand relative distance and relative rotation angle should be identical in all input images. Since the extrinsic parameters of the cameras from each viewpoint are different, predictions of the rotation angles and relative distances for each are coupled with the camera's rotation and translation predictions. Therefore, in each frame, we perform the same rotation and translation operations on the predicted joints of both hand to eliminate the effect of camera parameters. This process moves the root joint of the right hand to the coordinate origin and sets the rotation angle to zero, then denote the transformed left hand joints as $J^*$. The consistency loss is calculated:
\begin{equation}
    \label{eq_loss_consis}
    L_{consis} = \sum_{i=1}^{N_t}\sum_{k=1}^{N_t} ||J^*_i - J^*_k||^2
\end{equation}

\begin{figure*}[ht]
	\centering
	\includegraphics[width=1.0\textwidth]{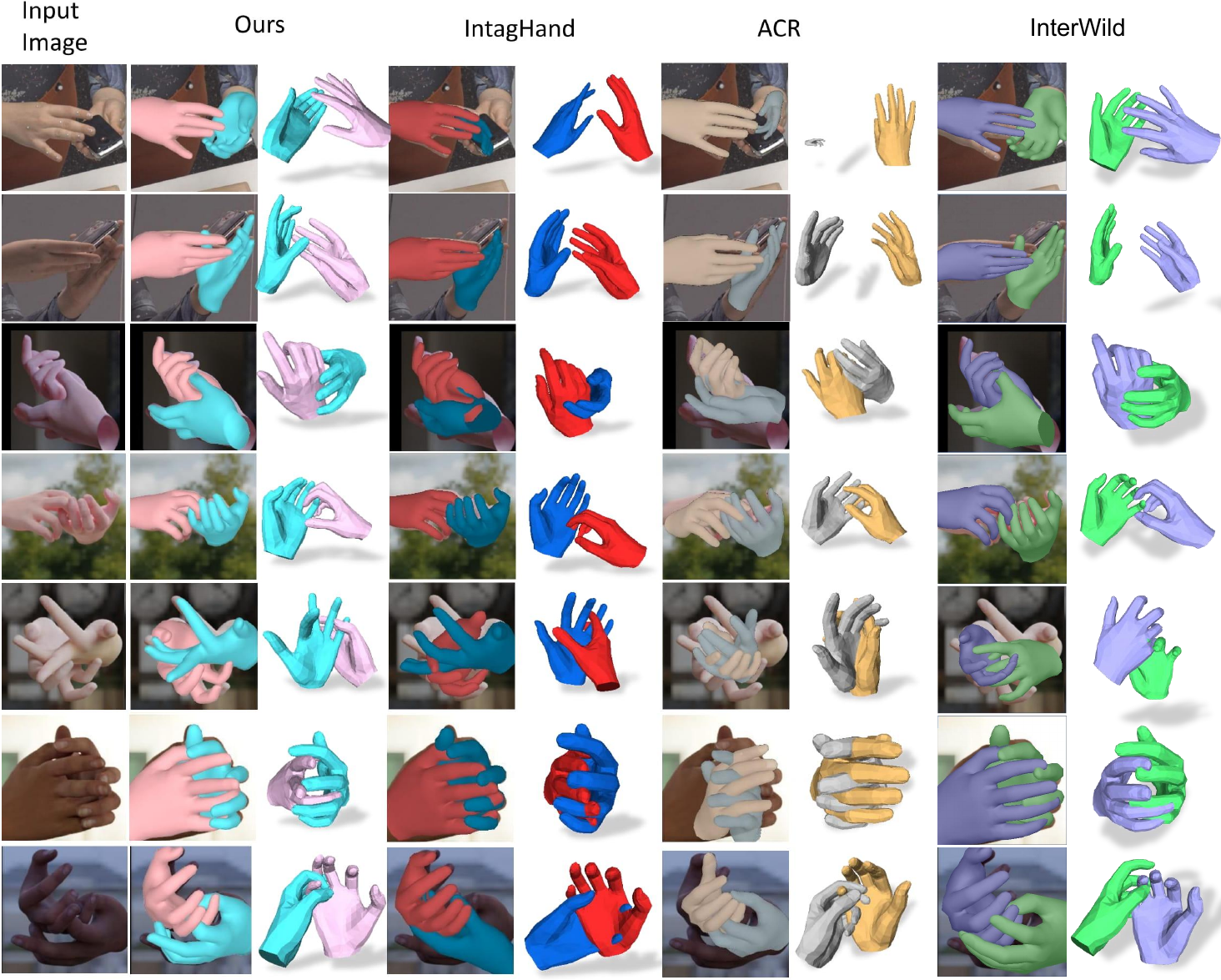}
    \vspace{5mm}
	\caption{ {Qualitative Comparison. We compare OmniHands with state-of-the-art methods on in-the-wild datasets ARCTIC\cite{arctic} and RenderIH\cite{renderIH}. ARCTIC provides two-hand object interacting images, and RenderIH provides interactive hands under different lighting conditions. We have demonstrated the results from both the camera perspective and an arbitrary perspective. The results show that our method outperforms the state-of-the-art methods in complex real-world environments. }}
    \Description{A Figure.}
	\label{fig_qua_result}
\end{figure*}

\begin{figure*}[ht]
	\centering
	\includegraphics[width=1.0\textwidth]{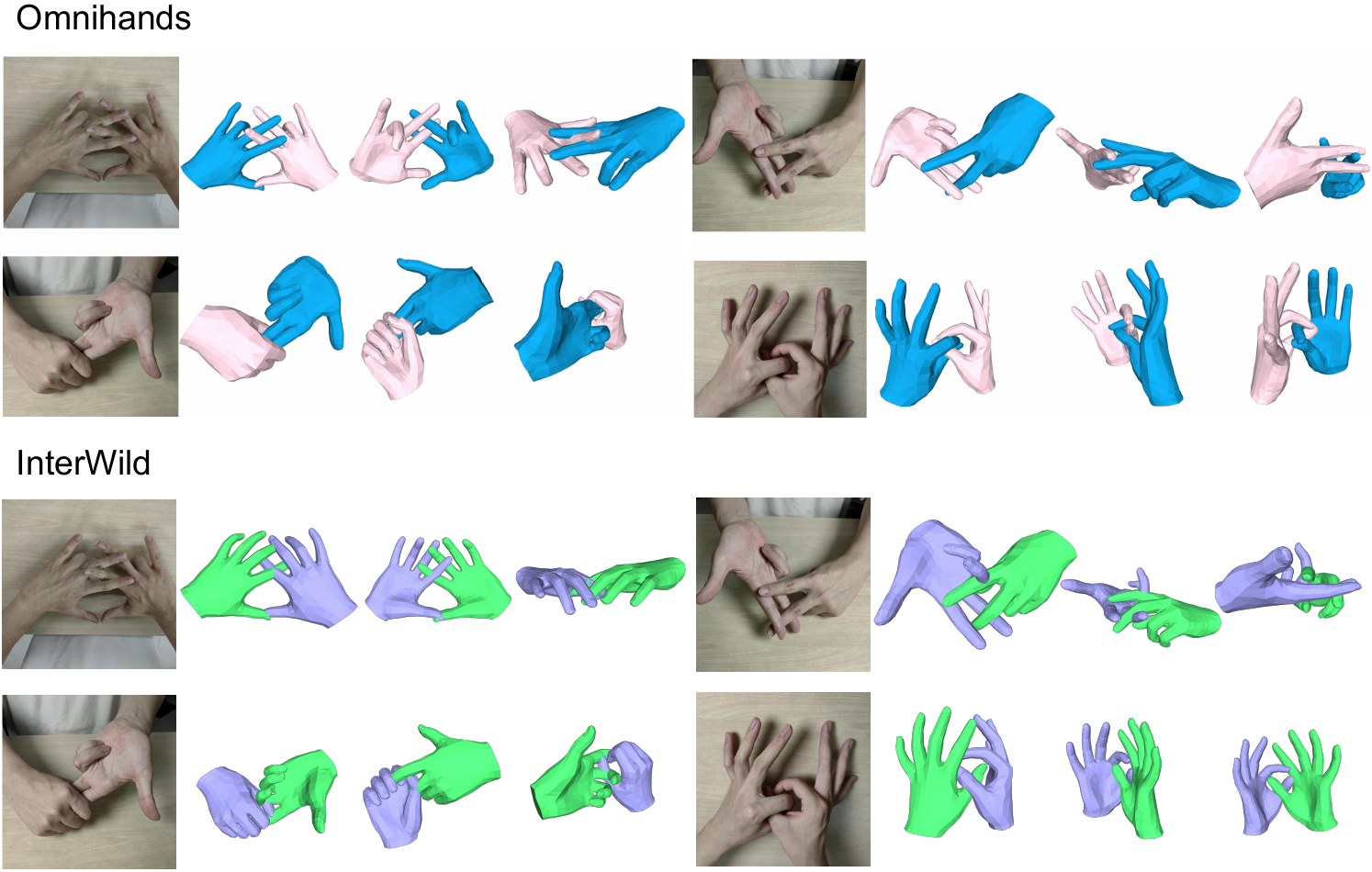}
    \vspace{-2mm}
	\caption{Qualitative Results of in-the-wild hard cases. We show our model's results on complex cross-hand interactions in realistic scenarios to demonstrate its performance. For a detailed perception, views of multiple perspectives are demonstrated.} 
    \Description{A Figure.}
	\label{fig_multi}
\end{figure*}

\section{Experiments}
\subsection{Implementation details}
We implement our model based on PyTorch~\cite{paszke2019pytorch}. We use a pretrained Vision Transformer following a huge design (ViT-h) with 32 layers as the backbone to encode image features. The resolution of input image is $256\times 192$ and the size of the feature map is $H=16, W=12, C=1280$. We use transformer encoders and decoders in our feature fusion modules. We train our model for 1,000,000 steps using AdamW optimizer with a linear learning rate decay schedule from an initial learning rate $1\times10^{-5}$, and the decay rate is 0.7 for every 200,000. The training set is mixed of Interhand2.6m~\cite{Moon_2020_ECCV_InterHand2.6M}, Re:Interhand~\cite{relighted} and Single Hand Datasets, we use a random sampling strategy with sampling rate 40\%, 40\%, 20\%. When taking temporal sequence as input, the length of an image sequence is set to 9 with a gap of 5 frames to sample images in a consecutive video sequence. 
\subsection{Datasets}
We use a training dataset with a mix up of interacting hand datasets and single-hand datasets. The interacting hand datasets are used to optimize the hand relations in cases that two hands are close. The single-hand datasets are used to optimize the accuracy of single-hand mesh prediction in cases where two hands are far away.

\textbf{Interhand2.6M} \cite{Moon_2020_ECCV_InterHand2.6M}is the first publicly available dataset of sequences of interacting hands with accurate annotations, which is widely used for training and testing in two-hand reconstruction methods. InterHand2.6M in 30 fps is for video-related research and the 5 fps version is a subset of the 30 fps version which removes the redundancy for single image research. In our method, we make use of 30 fps and 5 fps interacting two-hand (IH) data with human and machine (H+M) annotations and filter invalid data following \cite{intaghand}. We ultimately utilize 1.6M images for training and 1.2M images for testing.

\textbf{Re:InterHand} \cite{relighted} has images with realistic and diverse appearances along with accurate GT 3D interacting hands. It has 738K video-based images in 3rd-person viewpoints. We take the whole dataset for training. 

\textbf{DexYCB} \cite{DexYCB} contains sequences of single hands interacting with objects. It consists of 1000 sequences with 582K images. Although it's not a two-hand dataset, we use it as part of the training set and testing set to enhance our model's performance in scenarios with object occlusion and motion blur.

\textbf{Static Single Hand Datasets.} To enhance the performance in wild videos, we use single-hand datasets that provide 3D hand annotations for training. Following the previous work~\cite{hamer}, we use the datasets with a combination of FreiHAND~\cite{freihand}, HO3D~\cite{ho3d}, MTC~\cite{mtc}, RHD~\cite{rhd}, H2O3D~\cite{ho3d}, DEX YCB~\cite{DexYCB}, COCO WholeBody~\cite{COCOWhole}, Halpe~\cite{alphapose} and MPII NZSL~\cite{mpii}.  {We simulate a pair of widely separated hands by randomly combining two single-hand images and assigning them two non-overlapping bounding boxes.}

\textbf{In-the-wild Datasets.} We use in-the-wild datasets HIC \cite{hic}, ARCTIC \cite{arctic} and RenderIH \cite{renderIH} to evaluate the performance on various scenarios. HIC provides diverse hand-hand interacting and object-hand interacting sequences and contains 3D GT meshes of both hands. ARCTIC is a multi-view sequential dataset of two-hand interacting with daily use objects, involving occlusions and detailed interactive actions. RenderIH provides realistically rendered hands with a variety of poses, textures, backgrounds, and illuminations, which can simulate various scenarios in reality. For all in-the-wild datasets, we only use them for testing to evaluate the performance on unseen scenarios.

\subsection{Metrics}
We adopt Mean Per Joint Position Error (MPJPE) and Mean Per Vertex Position Error (MPVPE) and Area Under Curve (AUC) scores to measure the accuracy of the pose and shape of the estimated meshes, and Mean Relative-Root Position Error (MRRPE) to measure the accuracy of relative positions of two hands. For mean position error and AUC we get results under both Root-relative analysis (Ra) and Procrustes analysis (Pa). Following the prior method \cite{intaghand}, we perform root joint alignment and skeleton scaling when evaluating MPJPE. In addition, we use acceleration error Accel\_E to evaluate the temporal consistency, following VIBE\cite{vibe}. \textcolor{black}{To better evaluate the accuracy of two-hand interactions, we propose Mean Inter-Joint Error (MIJE) as an evaluation metric.}

\subsection{Qualitative Results}
 {For monocular inputs, our qualitative comparisons are shown in Figure \ref{fig_qua_result}. We demonstrate results of in-the-wild datasets ARCTIC~\cite{arctic} and RenderIH~\cite{renderIH}, which are unseen in the training phase. We compare with IntagHand\cite{intaghand}, ACR \cite{acr} and InterWild\cite{inthewild} in camera perspective and an arbitrary perspective to show more details. We show more results on realistic complex cross-hand interacting samples in Figure \ref{fig_multi}, comparing with InterWild. The results demonstrate the stability of OmniHands in real-world scenarios and in complicated interaction cases.} 

\begin{figure*}[ht]
	\centering
	\includegraphics[width=0.9\textwidth]{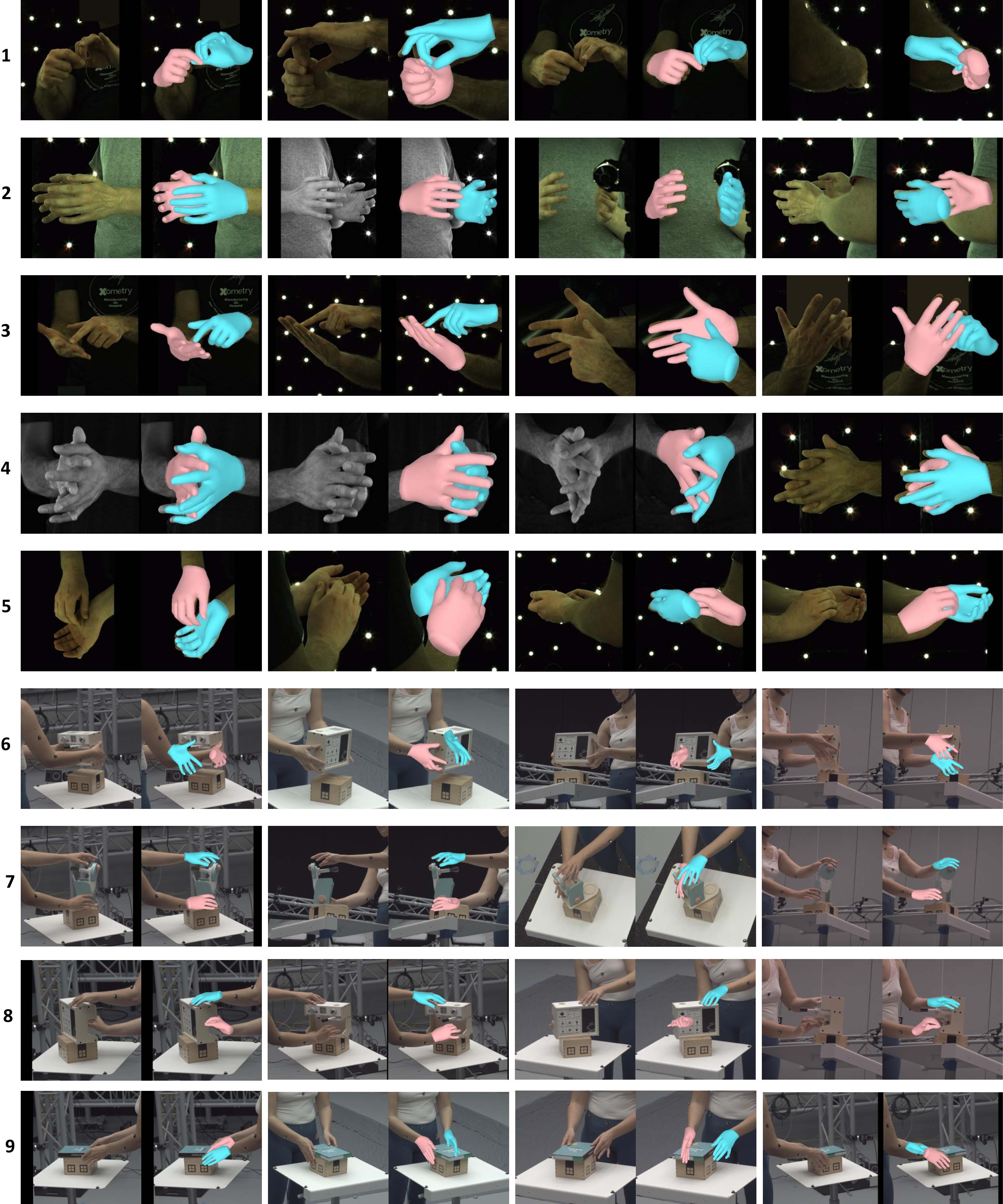}
    \vspace{-3mm}
    \caption{ {The qualitative results of OmniHands with multi-view inputs. We present the results on Interhand2.6m~\cite{Moon_2020_ECCV_InterHand2.6M} at row 1 to row 5 and results on Arctic~\cite{arctic} at row 6 to row 9. In each row, we present pairs of input and output images from 4 different camera views. The images on the left side of each pair serve as the multi-view sequence input. }
    }
    \Description{A figure.}
	\label{fig_mul_qua}
\end{figure*}

For multi-view inputs, the qualitative results are shown in Figure \ref{fig_mul_qua}, where 4 camera views are selected as demonstration. In every perspective, the hands are obscured to varying degrees. It can be observed that by integrating multi-view information, the hand interaction information is mutually supplemented, allowing successful reconstruction of the hand mesh corresponding to the respective view, even when the hands are completely occluded in some perspectives. It is worth noting that due to the absence of camera parameter inputs, our model needs to predict the camera angle solely based on visual information and determine the hand orientation in the presence of occlusion, which is a non-trivial achievement.

\subsection{Comparison with state-of-the-art methods}

To demonstrate the versatility and stability of OmniHands across various scenarios,  {we conducted comparative analyses with state-of-the-art methods in multiple domains, including temporal two-hand evaluation, frame-based two-hand evaluation, temporal single hand evaluation, frame-based single hand evaluation and in-the-wild dataset evaluation.}

\textbf{Comparison with Temporal Two Hand Methods.}
In Table \ref{tab_ith_30fps} we compare the proposed OmniHands with previous temporal methods on Interhand2.6m 30fps~\cite{Moon_2020_ECCV_InterHand2.6M}. To demonstrate the benefits of using large mixed training datasets for the model, as well as the impact of using temporal features on model accuracy, we trained four groups of models based on whether additional training data and temporal features were used. For the training datasets, we use Interhand2.6m~\cite{Moon_2020_ECCV_InterHand2.6M}, Re:Interhand~\cite{relighted}, and Single Hand Datasets mixed up as our training set for a better model generalization. To ensure a fair comparison with other two-hand reconstruction methods, which are trained only on Interhand2.6m, we use the same training set to train our model and the results are also shown marked with \dag. As previously mentioned, OmniHands can take sequential frames or single frame as input, to figure out the impact of using temporal features, we report and compare the results of OmniHands with single frame and sequential frames as input, denoting single-frame version with \S. The results show that our method far outperforms the existing two-handed temporal models, with 23.34\%(7.37 mm $vs.$ 9.09 mm) improvement in MPJPE and 24.05\% (2.81 mm/$s^2$ $vs.$ 3.70 mm/$s^2$) improvement in Accel\_E. Comparing the results of whether using temporal features, it can be observed that using temporal feature results in better stability between frames and improvement of key points accuracy. The Accel\_E improves 18.86\% (2.81 mm/$s^2$ $vs.$ 3.34 mm/$s^2$) and 17.55\% (3.02 mm/$s^2$ $vs.$ 3.55 mm/$s^2$) respectively, and there is also a notable improvement on MPJPE and MPVPE. By changing the training set, we can find that using extra training data can obviously improve the accuracy and stability. 

\textbf{Comparison with Frame-based Two Hand Methods.}
 {To compare with state-of-the-art frame-based methods and demonstrate the performance in single frame input scenarios, we conduct experiments at Interhand2.6m 5fps~\cite{Moon_2020_ECCV_InterHand2.6M} by setting the sequence length to 1.} As can be observed at Table \ref{tab_ith_5fps}, even though we drop the temporal knowledge, our method still keep high performance and outperforms previous methods that were specially designed for single frame input scenarios, achieving 7.49 mm, 7.72 mm and 24.58 mm in MPJPE, MPVPE and MRRPE.  {The improvement of our model becomes more significant when trained on the full dataset, achieving 6.77 mm, 6.96 mm and 22.01 mm in MPJPE, MPVPE and MRRPE.} We also report the results of multi-view frame inputs, without any camera parameter, our model achieves 5.93 mm, 6.20 mm and 14.08 mm in MPJPE, MPVPE and MRRPE, which significantly surpasses the accuracy of monocular estimation.

\begin{table}[htbp]
\caption{ {Quantitative comparisons with frame-based and temporal methods on Interhand2.6m 30fps dataset. The sign $\dag$ stands for using only Interhand2.6m 30fps as training set, and $\S$ stands for using frame-based input.}}
\label{tab_ith_30fps}
\begin{center}
\resizebox{1.0\linewidth}{!}{
\begin{tabular}{lccc}
\toprule
Methods  &Accel\_E$\downarrow$    & MPJPE$\downarrow$ & MPVPE $\downarrow$                  \\
\hline
HPS \cite{boukhayma20193d}      & 8.14 & 19.40 & 19.78 \\
ITH-3D\cite{ith3d}    & 6.01 & 14.45 & 14.74 \\
SeqHand\cite{seqhand}   & 6.69 & 18.20 & 19.46 \\
Zhao et. al.\cite{estc}      & 3.70 & 13.45 & 13.91 \\
Intaghand\cite{intaghand} & 4.40 & 9.26  & 9.71  \\
DIR\cite{decoupled}       & 8.92 & 9.09  & 9.43  \\
Ours $\dag$ $\S$  & 3.55 & 7.99  & 8.15  \\
Ours $\dag$  & 3.02 & 7.59  & 7.79  \\
Ours $\S$  & 3.34 & 7.80  & 7.95  \\
Ours   & \textbf{2.81} & \textbf{7.37}  & \textbf{7.58}  \\
\bottomrule
\end{tabular}}
\end{center}
\vspace{-0.2cm}
\end{table}

\begin{table}[htbp]
\caption{ {Performance comparisons on the Interhand2.6m 5fps dataset. We compare with the most recent methods. 'Ours$\dag$' was trained on Interhand2.6m 5fps. 'Ours' was trained on mixed dataset. 'Ours$\diamond$' takes multi-view images as input and was trained on Interhand2.6m 5fps. }}
\label{tab_ith_5fps}
\begin{center}
\resizebox{1.0\linewidth}{!}{
\begin{tabular}{lccc}
\toprule
Methods      & MPJPE$\downarrow$ & MPVPE$\downarrow$ & MRRPE$\downarrow$                  \\
\hline
InterWild\cite{inthewild}        & 14.76 & 13.24 & 26.25  \\

IntagHand\cite{intaghand}            & 8.79  & 9.02  &   27.74     \\
MEMAHand\cite{memahand}         & 8.65  & 8.89  &   -     \\
ACR\cite{acr}              & 8.41  & 8.53  &   -     \\
Zuo et.al\cite{reconprior}       & 8.34  & 8.51  &   -     \\ 
DIR\cite{decoupled}         & 7.51 & 7.72 & 28.98  \\
Ours$\dag$        & 7.49 & 7.72 & 24.58 \\
Ours     & \textbf{6.77} & \textbf{6.96} & \textbf{22.01} \\
Ours $\diamond$        & 5.93 & 6.20 & 14.08 \\

\bottomrule
\end{tabular}}
\end{center}
\vspace{-0.2cm}
\end{table}

\begin{table*}[!htbp]
\caption{ {Multi-view single hand recovery results on Dex-YCB, comparing with other single hand multi-view methods. The model reported was trained on Dex-YCB.}
}
\vspace{-2mm}
\label{tab_muldex}
\begin{center}
\resizebox{1\textwidth}{!}{
\begin{tabular}{lllllll}
\toprule
 Method    & Ra-MPJPE$\downarrow$ & Pa-MPJPE$\downarrow$ & Ra-JAUC$\uparrow$ & Ra-MPVPE$\downarrow$ & Pa-MPVPE$\downarrow$ & Ra-VAUC$\uparrow$ \\
     \hline
MVP\cite{MVPzhang2021direct}  & 6.23    & 4.26    & 0.69 & 9.77    & 8.14    & 0.53 \\
PE-Mesh-TR\cite{yang2023poem} & 7.49 & 4.76 & 0.64 & 7.41 & 4.70 & 0.64 \\
FTL-Mesh-TR\cite{yang2023poem} & 8.66 & 5.51 &  0.59 & 8.75 & 5.75 &  0.59 \\ 
POEM\cite{yang2023poem} & \textbf{6.06}    & 3.93    & 0.68 & \textbf{6.13}    & 4.00    & 0.70 \\
Multi-view Fit.\cite{yang2023poem} & 7.33    & 5.29    & 0.65 & 7.22    & 5.19    & 0.65 \\
Ours & 6.44    & \textbf{3.18 }   & \textbf{0.87} & 6.42    & \textbf{3.26}    & \textbf{0.87} \\
\bottomrule
\end{tabular}}
\end{center}
\vspace{-0.2cm}
\end{table*}

\begin{table*}[!htbp]
\caption{ {Comparison of two-hand reconstruction methods on in-the-wild datasets. Arctic, HIC and RenderIH are taken as test set, which are unseen during training. Our method marked with $\dag$ is trained on only Interhand2.6m, and the unmarked one is trained on mixed training set.} 
}
\vspace{-3mm}
\label{tab_wild}
\begin{center}
\resizebox{1.0\textwidth}{!}{
\begin{tabular}{c|c|lllll|llll}
\toprule
\multirow{2}{*}{Dataset}  & \multicolumn{1}
{c}{\multirow{2}{*}{Method}} & \multicolumn{5}{|c|}{Joints}            & \multicolumn{4}{c}{Vertices}          \\
\cline{3-11}

                          & \multicolumn{1}{c|}{}                        & Ra-MPJPE$\downarrow$ & Pa-MPJPE$\downarrow$ & Ra-JAUC$\uparrow$ & Pa-JAUC$\uparrow$ & MRRPE$\downarrow$ & Ra-MPVPE$\downarrow$ & Pa-MPVPE$\downarrow$ & Ra-VAUC$\uparrow$ & Pa-VAUC$\uparrow$  \\
                          \hline
\multirow{5}{*}{Arctic\cite{arctic}}   & IntagHand\cite{intaghand}                                       & 24.88    & 12.02    & 0.57   & 0.76 &203.04  & 26.30    & 12.52    & 0.55   & 0.75   \\
                          & InterWild\cite{inthewild}                                    & 22.74    & 11.31    & 0.59   & 0.72 & 143.19   & 23.92    & 11.74    & 0.56   & 0.71   \\
                          & DIR\cite{decoupled}                                         & 26.63    & 11.97    & 0.55   & 0.76 & 253.49   & 28.33    & 12.39    & 0.52   & 0.75   \\
                          & ours \dag                    & 14.21    & 9.22     & 0.69   & 0.78 & 112.81   & 15.12    & 9.76     & 0.63   & 0.81   \\
                          & ours                                        & \textbf{13.57}    & \textbf{7.98}     & \textbf{0.74}   & \textbf{0.84} &\textbf{105.43}  & \textbf{14.51}    & \textbf{8.75}     & \textbf{0.72}   & \textbf{0.82}   \\
                          \hline
\multirow{5}{*}{HIC\cite{hic}}      & IntagHand\cite{intaghand}                                       & 12.01    & 7.46     & 0.76   & 0.85 &36.46   & 12.77    & 7.82     & 0.74   & 0.84   \\
                          & InterWild\cite{inthewild}                                  & 11.23    & 6.62     & 0.79   & 0.86 &30.35   & 11.65    & 6.99     & 0.76   & 0.86   \\
                          & DIR \cite{decoupled}                                        & 11.90    & 7.12     & 0.76   & 0.85 &32.83  & 12.79    & 7.55     & 0.75   & 0.84   \\
                          & ours \dag                    & 9.47     & 5.55     & 0.81   & 0.88 &26.84   & 10.35    & 6.07     & 0.79   & 0.87   \\
                          & ours                                        & \textbf{8.66}     & \textbf{4.81}     & \textbf{0.82}   & \textbf{0.90} &\textbf{26.56}  & \textbf{9.31}     &\textbf{ 5.30}     & \textbf{0.81}   & \textbf{0.89}   \\
                          \hline
\multirow{5}{*}{RenderIH\cite{renderIH}} & IntagHand\cite{intaghand}                                       & 33.84    & 15.19    & 0.47   & 0.70 &80.28  & 35.17    & 15.34    & 0.45   & 0.69   \\
                          & InterWild  \cite{inthewild}                                 & 19.12    & 13.16    & 0.62   & 0.72  &56.10 & 19.74    & 12.77    & 0.61   & 0.73   \\
                          & DIR\cite{decoupled}                                         & 29.48    & 12.93    & 0.53   & 0.74 &78.53  & 30.93    & 12.89    & 0.51   & 0.74   \\
                          & ours \dag                    & 17.22    & 10.91    & 0.66   & 0.78 &45.94  & 18.01    & 11.11    & 0.64   & 0.77   \\
                          & ours                                        & \textbf{13.71}    & \textbf{9.17}     & \textbf{0.72}   & \textbf{0.81}  &\textbf{42.26} & \textbf{14.32}    & \textbf{9.48}     & \textbf{0.71}   & \textbf{0.81}  \\
                          \bottomrule
\end{tabular}}
\end{center}
\vspace{-0.2cm}
\end{table*}

\textbf{Comparison on Single hand Dataset.}
 {
Although our framework is primarily designed for interacting hands reconstruction, it also achieves performance comparable to state-of-the-art single-hand methods on the task of single-hand reconstruction.  We test OmniHands on 2 widely used benchmarks: Freihand~\cite{freihand} and HO3Dv2~\cite{ho3d}, , with the results presented in Table~\ref{tab_frei}, Table~\ref{tab_ho3d}. For a fair comparison, we adopt the same training settings as Hamba~\cite{dong2024hamba}. Specifically, the model in Table~\ref{tab_frei} is trained only using the FreiHAND dataset, and the model in Table~\ref{tab_ho3d} is trained on the same datasets as HaMeR~\cite{hamer}. 
}

\textcolor{black}{In the task of interacting-hand reconstruction, we primarily use MRRPE to evaluate the accuracy of the relative root positions, specifically the wrist positions, between the two hands as an indicator of interaction accuracy. However, such a numerical metric does not effectively capture the front-back ordering of the fingers between the hands, which can lead to noticeable differences in visual quality. Therefore, we propose Mean Inter-Joint Error (MIJE) to evaluate the hand interaction performance. The MIJE is computed as follows:}

\begin{equation}
    \mathbf{MIJE} = \frac{1}{|J_r|\cdot|J_l|}\sum_{j_r\in J_r}\sum_{j_l\in J_l} \|j_r-j_l\|_2
\end{equation}

\textcolor{black}{We evaluated this metric on InterHand2.6M and compared it with state-of-the-art methods, the results are reported in Table \ref{tab:MIJE}. As shown, our method achieves notably better performance in handling complex two-hand interaction poses. This advantage is also reflected in other qualitative results.}

\begin{table}[htbp]
\caption{\textcolor{black}{The Mean Inter-Joint Error (MIJE) metric, measured in millimeters, of our method compared to the SOTA in two-hand reconstruction on the InterHand2.6M dataset. Lower values indicate better performance.}
}
\label{tab:MIJE}
\begin{tabular}{cccc}
\toprule
Ours  & IntagHand & InterWild & DIR    \\
\hline
\textbf{26.56} & 32.36     & 58.51     & 73.40 \\
\bottomrule
\end{tabular}
\end{table}

\textbf{Comparison with Temporal Single Hand Methods.}
To further exploit the versatility of our method, we conduct experiments on temporal single hand dataset Dex-YCB~\cite{DexYCB}. We manage our input by flipping an input image as the opposite hand, and using far away bounding boxes to indicate the hands are not interacting. We report the results of Root-Aligned and Procrustes-aligned metrics. It can be observed in Table \ref{tab_dex} that our method outperforms all existing single hand temporal methods in all metrics. It is noteworthy that our lead in Root-Aligned metrics is greater than those in Procrustes-aligned metrics (+38.63\% $vs .$ +9.00\% in MPJPE, +12.70\% $vs .$ +0.89\% in JAUC), indicating that we can predict accurate shapes and rotation angles without the assistance of ground truth alignment.

\begin{table}[htbp]
\caption{Quantitative comparisons with other single hand reconstruction methods on Freihand dataset. }
\vspace{-3mm}
\label{tab_frei}
\begin{center}
\resizebox{\linewidth}{!}{
\begin{tabular}{lcccc}
\toprule
Method & PA-MPJPE $\downarrow$     & PA-MPVPE $\downarrow$    & F@5mm $\uparrow$ &  F@15mm  $\uparrow$\\
\hline
PointHMR\cite{kim2023pointhmr}                     & 6.1          & 6.6          & 0.720                              & 0.984                               \\
zhou et. al \cite{zhou2024simple}                  & 5.7          & 6.0          & 0.772                              & 0.986                               \\
HaMeR \cite{hamer}                       & 6.0          & 5.7          & 0.785                              & 0.990                                \\
Hamba  \cite{dong2024hamba}                      & 5.8          & \textbf{5.5} & \textbf{0.798}                     & 0.991                               \\
Ours                         & \textbf{5.2} & 5.7          & 0.779                              & \textbf{0.993}    \\
\bottomrule
\end{tabular}
}
\end{center}
\end{table}

\begin{table}[htbp]
\caption{Quantitative comparisons with other single hand reconstruction methods on HO3Dv2 dataset. }
\vspace{-3mm}
\label{tab_ho3d}
\begin{center}
\resizebox{\linewidth}{!}{
\begin{tabular}{lcccc}
\toprule
Method & PA-MPJPE $\downarrow$     & PA-MPVPE $\downarrow$    & F@5mm $\uparrow$ &  F@15mm  $\uparrow$\\
\hline
H2ONet   \cite{xu2023h2onet}                    & 8.5          & 8.6          & 0.966                              & 0.570                               \\
HandBooster   \cite{xu2024handbooster}               & 8.2          & 8.4          & 0.972                              & 0.585                               \\
HaMeR \cite{hamer}                       & 7.6          & 7.9          & 0.980                               & 0.635                               \\
Hamba  \cite{dong2024hamba}                      & 7.5          & \textbf{7.7} & 0.982                     & \textbf{0.648}                      \\
Ours                         & \textbf{6.9} & \textbf{7.7} & \textbf{0.985}                     & 0.616    \\
\bottomrule
\end{tabular}
}
\end{center}
\end{table}

\begin{table}[htbp]
\caption{ {Quantitative comparisons with other single hand temporal reconstruction methods on the Dex-YCB dataset. The model reported was trained on Dex-YCB}}
\vspace{-3mm}
\label{tab_dex}
\begin{center}
\resizebox{\linewidth}{!}{
\begin{tabular}{lcccc}
\toprule
Methods      & Ra-MAJPE$\downarrow$  & Ra-JAUC$\uparrow$  & Pa-MPJPE$\downarrow$  & Pa-JAUC$\uparrow$                  \\
\hline
HaMeR \cite{hamer} & 19.45& 68.4 & 6.09 & 87.8 \\
VIBE\cite{vibe}                                  & 16.95 & 67.5    & 6.43     & 87.1    \\
TCMR\cite{tcmr}                                  & 16.03 & 70.1    & 6.28     & 87.5    \\
Tu et al. \cite{s2handv}                               & 19.67 & 62.5    & 7.27     & 85.5    \\
MeshGraphormer\cite{meshgraphormer} &16.21 &69.1 &6.41 &87.2 \\
Deformer\cite{deformer}                              & 13.64 & 74.0    & 5.22     & 89.6    \\
Ours        & \textbf{8.37} & \textbf{83.4} & \textbf{4.75} & \textbf{90.4} \\

\bottomrule
\end{tabular}
}
\end{center}
\end{table}

\textbf{Quantitative Comparison on In-the-wild Datasets.}
We report the results on three in-the-wild datasets: Arctic~\cite{arctic}, HIC~\cite{hic} and RenderIH~\cite{renderIH}, to compare the performance at in-the-wild scenarios. Arctic contains images of two hand interacting with objects,  {including a large number of cases with hand-object occlusion and hand-hand occlusion.} HIC is a small dataset of two hand interaction sequences.  {And RenderIH contains images of complicated two hand interactions under various illumination conditions.} We only use samples in which both hands are visible. For fair comparison, temporal data or multi-view data are not used in these experiments, a single image is taken as input for each case. 

In Table \ref{tab_wild} we compare the results with IntagHand~\cite{intaghand}, InterWild~\cite{inthewild} and DIR~\cite{decoupled}. IntagHand~\cite{intaghand} and DIR~\cite{decoupled} are two open-source inter-hand reconstruction models that have demonstrated excellent performance in training and testing on Interhand2.6m~\cite{Moon_2020_ECCV_InterHand2.6M}. InterWild~\cite{inthewild} is a SOTA method specifically designed for two-hand recovery in the wild. To demonstrate the efficiency of using large training data, we report results of models trained under two training set.  {One of our reported models is trained solely on InterHand2.6M, using the same training set as other comparison methods to enable a fair quantitative comparison; this model is marked with $\dag$. The other unmarked model is trained on a mixed dataset comprising InterHand2.6M, Re:InterHand, and single hand datasets.}

In all three unseen datasets, our method achieves the best performance and leads in all metrics. It can also be found that by using large training set, our model can handle situations in more diverse scenarios, resulting in higher accuracy in all three testing sets.
HIC~\cite{hic} includes interactive gestures that are not present in Interhand2.6m. However, the scenes are relatively simple and the test samples are few, allowing other methods to also achieve good results. In HIC our method improves 16.33\% (8.66 mm $vs.$ 10.35 mm) on Ra-MPJPE and 29.47\% (4.81 mm $vs.$ 3.34 mm) on Pa-MPJPE. 
Arctic~\cite{arctic} includes a large number of images of two hands interacting with objects, which significantly differ from the data in Interhand2.6m. Additionally, there are many instances where parts of the hands are occluded, making the situation more complex. In this dataset, other methods performed poorly, whereas our method maintained robustness and achieved excellent results, improving 45.46\% (13.57 mm $vs.$ 24.88 mm) on Ra-MPJPE and 33.61\% (7.98 mm $vs.$ 12.02 mm) on Pa-MPJPE. 
In RenderIH~\cite{renderIH}, the diversity of lighting and backgrounds places high demands on the cross-scene robustness of the model. Our proposed OmniHands obtains 13.89\% (13.71 mm $vs.$ 15.92 mm) improvement on Ra-MPJPE and 13.00\% (9.17 mm $vs.$ 10.54 mm) on Pa-MPJPE.

\textbf{Comparison on Single hand Multi-view Dataset.}
To demonstrate the performance of our model when taking multi-view images as input, in Table \ref{tab_muldex} we compare with other multi-view single hand methods on Dex-YCB~\cite{DexYCB}.  {For a fair comparison, our model is trained on Dex-YCB.} We also compare with Multi-view pose estimation reported in \cite{yang2023poem}. It's worth mentioning that other methods require camera parameters as input, which makes it easier for them to align different views but loses flexibility. Our method, on the other hand, does not need camera parameters as input. It only requires multiple images from different viewpoints, eliminating the tedious camera calibration steps while achieving higher accuracy than other methods.

\subsection{Ablation Study}
We evaluate the importance of each module in OmniHands by ablation experiments in temporal input conditions, using temporal dataset Interhand2.6m 30fps as training set and testing set. Moreover, we also conduct evaluations on the impact of camera numbers when taking multi-view images as input, using Interhand2.6m 5fps multi-view dataset.

\subsubsection{Module Ablation}
\textbf{ViT Baseline.} We trained a baseline using ViT and a transformer decoder on the full training set. Position map embedding, cross-hand feature interaction and temporal/multi-view interaction are not employed on the baseline, so the Bi-manual Decoder, Collective Decoder, and Individual Encoder are removed. We show the results in Table \ref{tab_ablation}. Although ViT has high performance in detecting single hand, the occlusion caused by hands interaction reduces its predictive ability. Besides, without cross-hand knowledge, it separately predicts the positions of two hands, causing huge errors in relative position.

\begin{figure}[h]
  \centering
  \includegraphics[width=\linewidth]{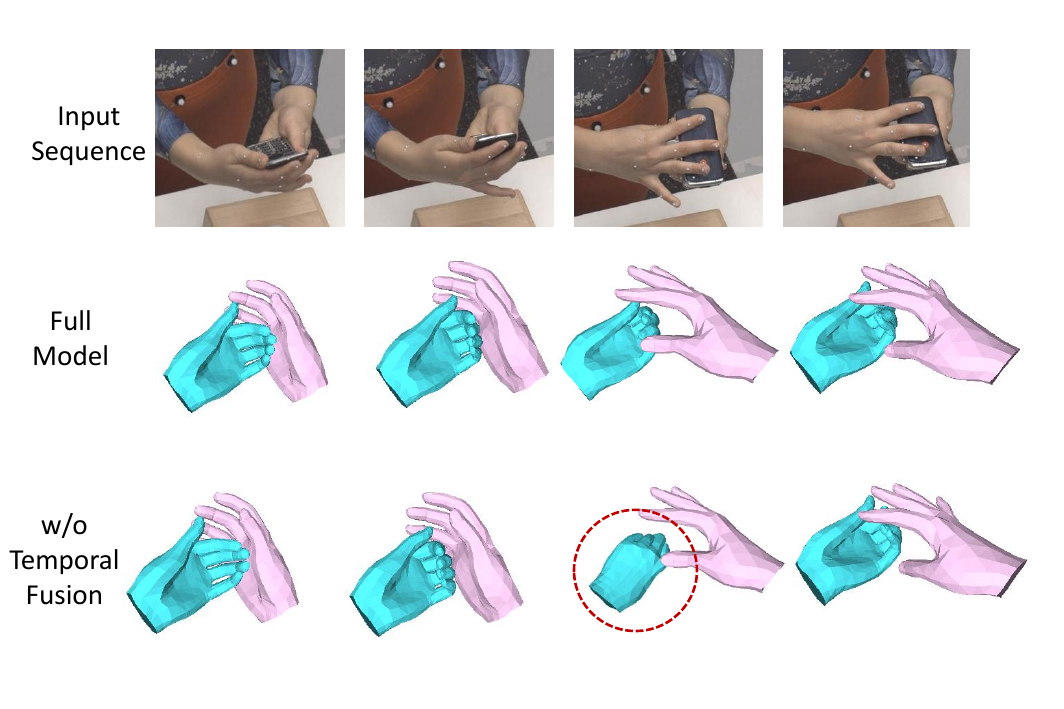}
  \caption{Ablation study on occlusion case. 'Full Model' is our model with all modules. 'w/o Temporal Fusion' means that the temporal integrating module is removed.}
  \Description{A figure.}
  \label{fig_abl_temp}
\end{figure}

\textbf{Effectiveness of Cross-hand Information.} We experimented in adding cross-hand information in two ways: (1) Input holistic image including two hands(+Cross H). (2) Input separate images, each image contains a single hand(+Cross S). In both way we add Bimanual Decoder module to integrate cross-hand information, the results show that the last way achieves better accuracy. The main reason is that by separating two hands as input, the visual encoder focuses on extracting single hand features, allowing the ViT to learn more hand details when learning on large scale datasets.

\begin{figure}[h]
  \centering
  \includegraphics[width=\linewidth]{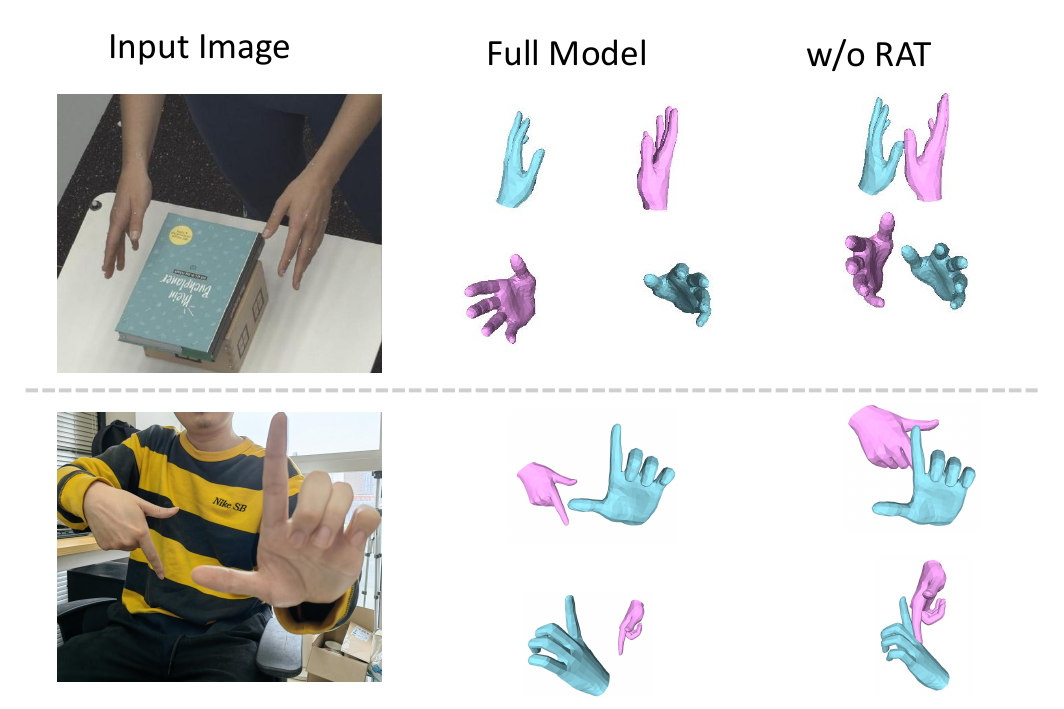}
  \caption{Ablation study of RAT. Two views of predicted hand meshes are visualized to demonstrate details of relative position. 'Full Model' is our model with all modules. 'w/o RAT' means that the position embedding is removed.}
  \Description{A figure.}
  \label{fig_abl_map}
\end{figure}

\textbf{Effectiveness of Relation-aware Tokenization.} 
By removing position token map embedding in RAT(+Cross S w/o RAT), the model lost global position information during inference, which means that the model could only infer two-hand relative positions from visual information. In InterHand2.6m~\cite{Moon_2020_ECCV_InterHand2.6M}, the two hands are relatively close and the 2D scales of each hand are similar. Therefore, even without a global position information, the approximate positions of both hands can be inferred from the input image. We test cases when two hands are relatively far away and cases when 2D scales of each hand are different. As shown in Table ~\ref{fig_abl_map}, without position token map, the model performs poor in predicting two-hand relative position.

\textbf{Effectiveness of Temporal Fusion.}
By adding Collective Decoder for temporal information interacting(+Cross S + Seq), our method can make sequential consistent prediction therefore infer reasonable poses encountering occlusions, and achieves improvement of 0.32 mm in MPJPE. In Figure \ref{fig_abl_temp} we visualize the predicted mesh sequences in the case of severe occlusion. It shows that the left hand is entirely occluded, the model without temporal fusion makes erroneous prediction. Our full model predicts a consistent sequence with assistant information from the past and the future.

\subsubsection{Ablation on training data.}
 {
In this section, we conduct ablation studies to evaluate the impact of scaling up training data. We benchmark our method on three datasets: InterHand2.6m~\cite{Moon_2020_ECCV_InterHand2.6M}, HIC~\cite{hic}, and Arctic~\cite{arctic}, and compare against two SOTA baselines: IntagHand~\cite{intaghand} and InterWild~\cite{inthewild}. The results are presented in Table~\ref{tab_bal_data}. For each method, we train the models on both the InterHand2.6M 5fps dataset and mixed datasets. Specifically, the mixed datasets for OmniHands include InterHand2.6M, Re:InterHand, and Static Single Hand Datasets. The static single hand datasets simulate a pair of widely separated hands by randomly combining two single-hand images. However, such input is not suitable for IntagHand or InterWild, as their input requires images that contain both hands. Therefore, for IntagHand and InterWild, the mixed datasets comprise InterHand2.6M and Re:InterHand. Concerning the mixed datasets, while OmniHands leverages additional single-hand datasets during training, all methods are trained on the same interacting-hand datasets. As the prediction of relative hand positions is learned from the interacting-hand data, the MRRPE metric serves as a valid indicator for comparing the interacting-hand reconstruction performance among the methods.}

 {
The experimental results demonstrate that OmniHands benefits from the increased data diversity, exhibiting improved performance when trained with the hybrid dataset. In contrast, IntagHand suffers performance degradation with additional training data. This is likely due to its architectural limitations, which prevent it from effectively generalizing under increased data scale and complexity, such as varying scenes and lighting conditions, ultimately undermining its reliance on strong priors. Overall, these results validate that OmniHands consistently outperforms SOTA methods, regardless of whether the model is trained on a single-domain dataset or a more diverse hybrid dataset.
}

\subsubsection{Camera view Ablation}
In multi-view reconstruction, the number of input camera views is important to the resulting accuracy. We tried to figure out the impact of camera views, and the experimental results are shown in Table \ref{tab_view_num}. We divided the experimental results into two groups based on whether mixed training data was used. 'I' indicates the use of only Interhand2.6m\cite{Moon_2020_ECCV_InterHand2.6M} as the training data, while 'I+R+A' represents the use of a mixed training dataset comprising Interhand2.6m, Re:Interhand\cite{relighted}, and Arctic\cite{arctic}. 

From the results presented in the table, it can be observed that simply switching from monocular input to 2-camera multi-view input significantly improves our prediction accuracy, and this improvement is even more pronounced when training with a mixed dataset. As the number of cameras increases from 2 to 16, the accuracy metrics improve accordingly regardless of the training set. This aligns with the intuitive understanding that more perspective leads to more accurate predictions. This demonstrates that our FIR module effectively utilizes 4D data to enhance prediction results and shows significant flexibility in handling 4D data of varying lengths. Additionally, by comparing the results of 'I' and 'I+R+A' in the table, it is evident that the accuracy with the mixed training dataset is significantly higher than with a single training dataset, irrespective of the number of cameras. This highlights our model's capability to handle large-scale mixed data efficiently.

\begin{table}[htbp]
\caption{Ablation studies of the RAT and FIR modules on InterHand2.6M. 
}
\vspace{-3mm}
\label{tab_ablation}
\begin{center}
\resizebox{1.0\linewidth}{!}{
\begin{tabular}{lccc}
\toprule
Methods      & MPJPE$\downarrow$ & MPVPE$\downarrow$ & MRRPE$\downarrow$                  \\
\hline
Baseline                            & 10.94      & 11.35      &  137.53     \\
+Cross H                           & 8.78  & 9.01  & 28.65 \\
+Cross S            & 7.69  & 7.92  & 24.79 \\
+Cross S w/o RAT         & 7.81  & 8.00  & 25.81 \\
+Cross S + Seq (Ours)           & \textbf{7.37}  & \textbf{7.58}  & \textbf{24.48} \\

\bottomrule
\end{tabular}}
\end{center}
\vspace{-0.2cm}
\end{table}

\begin{table}[htbp]
\caption{Results on Interhand2.6m using different number of camera views. "I" means only using Interhand2.6m training set, and "I+R+A" means using Interhand2.6m, Re:Interhand and Arctic as training set. 
}
\vspace{-3mm}
\label{tab_view_num}
\begin{center}
\resizebox{1.0\linewidth}{!}{
\begin{tabular}{l|l|lll}
\toprule
Train & cameras & MPJPE$\downarrow$ & MPVPE$\downarrow$ & MRRPE$\downarrow$  \\
\hline
\multirow{5}{*}{I}          & 2     & 7.02 & 7.28 & 20.59 \\
          & 4     & 6.88 & 7.13 & 18.90 \\
          & 8     & 6.79 & 7.05 & 17.91 \\
          & 12    & 6.65 & 6.93 & 16.86 \\
          & 16    & 6.49 & 6.76 & 17.08 \\
          \hline
\multirow{5}{*}{I+R+A}          & 2     & 6.51 & 6.74 & 17.72 \\
         & 4     & 6.28 & 6.55 & 16.49 \\
          & 8     & 6.21 & 6.47 & 16.26 \\
          & 12    & 6.14 & 6.42 & 14.15 \\
         & 16    & 5.93 & 6.20 & 14.08 \\
\bottomrule
\end{tabular}}

\end{center}
\vspace{-0.2cm}
\end{table}

\begin{table}[!htbp]
\caption{Albation study on training dataset. Models denoted with '-i' is trained on Interhand2.6m. Models denoted with '-m' is trained on mixed datasets.
}
\vspace{-3mm}
\label{tab_bal_data}
\begin{center}
\resizebox{1.0\linewidth}{!}{
\begin{tabular}{c|c|ccc}
\toprule
\multicolumn{1}{l}{Dataset}   & \multicolumn{1}{|l|}{Methods} & MPJPE          & MPVPE          & MRRPE           \\ 
\hline
\multirow{6}{*}{Inter2.6m 5fps} & IntagHand-i                 & 8.79           & 9.02           & 27.74           \\
                                & IntagHand-m                 & 10.55          & 10.98          & 30.04           \\
                                & InterWild-i                 & 14.76          & 13.24          & 26.25           \\
                                & InterWild-m                 & 14.45          & 13.01          & 25.68           \\
                                & Ours-i                   & 7.49           & 7.72           & 24.58           \\
                                & Ours-m                   & \textbf{6.77}  & \textbf{6.96}  & \textbf{22.01}  \\
\hline
\multirow{6}{*}{HIC}            & IntagHand-i                 & 12.06          & 12.77          & 36.46           \\
                                & IntagHand-m                 & 12.54          & 13.32          & 35.41           \\
                                & InterWild-i                 & 11.23          & 11.65          & 30.35           \\
                                & InterWild-m                 & 11.13          & 11.54          & 30.59           \\
                                & Ours-i                   & 9.47           & 10.35          & 26.84           \\
                                & Ours-m                   & \textbf{8.66}  & \textbf{9.31}  & \textbf{26.56}  \\
\hline
\multirow{6}{*}{Arctic}         & IntagHand-i                 & 24.88          & 26.30          & 203.14          \\
                                & IntagHand-m                 & 27.64          & 32.78          & 225.49          \\
                                & InterWild-i                 & 22.74          & 23.92          & 143.19          \\
                                & InterWild-m                 & 21.85          & 22.98          & 134.69          \\
                                & Ours-i                   & 14.21          & 15.12          & 112.81          \\
                                & Ours-m                   & \textbf{13.57} & \textbf{14.51} & \textbf{105.43} \\
                                \bottomrule
\end{tabular}}
\end{center}
\vspace{-0.2cm}
\end{table}


\section{Limitaions }
 {Although OmniHands has demonstrated high accuracy in offline video-based reconstruction, there remain several challenges that currently hinder its deployment for real-time interacting-hand reconstruction.
Our method utilizes ViT-h as the backbone, resulting in higher computational demands compared to CNN-based methods, which is a common issue with transformer-based models, as seen in HaMeR~\cite{hamer} and Hamba~\cite{dong2024hamba}. During inference, running on a single A800 GPU with Python code, our model achieves 16.05 FPS, indicating that it is not suitable for real-time rendering, which requires an efficiency of at least 25 to 30 FPS. Additionally, since our approach relies on external detection models to localize hands and regress bounding boxes, the overall inference speed is also limited by the performance of these detection models. Another challenge in real-time reconstruction arises from the fact that we predict the middle frame, which introduces a minimum delay of 4 frames in a live video stream, as the model requires access to 4 future frames as input. We will focus on accelerating our model by reimplementing it in C++ instead of Python and optimizing the computation process to enable real-time rendering.}

\section{Conclusions}

In this paper, we presented OmniHands, a robust solution for recovering interactive hand motions and their relative movements from monocular inputs. OmniHands is a transformer-based method with novel tokenization and feature fusion strategies. To this end, we introduced the Relation-aware Two-Hand Tokenization (RAT) method, which embeds positional relation information into hand tokens. This allows our network to handle both single-hand and two-hand inputs, leveraging their relative positions to reconstruct detailed hand interactions in real-world scenarios. Moreover, this tokenization approach enhances feature fusion by indicating the relative positions of the hands. To further this goal, we developed a 4D Interaction Reasoning (FIR) module that fuses hand tokens in 4D with attention, decoding them into 3D hand meshes and their temporal movements. The effectiveness of OmniHands is demonstrated through extensive evaluations on benchmark datasets, with results from in-the-wild videos and real-world scenarios showcasing its superior performance in interactive hand reconstruction.

\bibliographystyle{ACM-Reference-Format}
\bibliography{reference, ref/ref_HHMR, ref/ref_intaghand, ref/ref_mvhand}

\end{document}